\documentclass[preprint, journal]{vgtc}                     

\onlineid{0}

\preprinttext{To appear in IEEE Transactions on Visualization and Computer Graphics.}

\vgtccategory{Research}
\usepackage{bbm}
\usepackage{mathtools}
\usepackage{svg}

\newcommand{\highlight}[1]{#1}
\newcommand{\myparagraph}[1]{\noindent{\textbf{#1}.}}

\vgtcpapertype{please specify}

\title{A Neural Field-Based Approach for\\View Computation \& Data Exploration in 3D Urban Environments}

\author{%
  \authororcid{Stefan Cobeli}{0000-0002-9277-1317}, \authororcid{Kazi Shahrukh Omar}{0000-0003-3182-6553}, Rodrigo Valença, \authororcid{Nivan Ferreira}{0000-0001-6631-4609}, and \authororcid{Fabio Miranda}{0000-0001-8612-5805}
}

\authorfooter{
  \item
  	Stefan Cobeli, Kazi Shahrukh Omar, and Fabio Miranda are with the University of Illinois Chicago.
  	E-mail: \{scobel2,komar3,fabiom\}@uic.edu
  \item
  	Rodrigo Valença and Nivan Ferreira are with the Universidade Federal de Pernambuco. E-mail:
    \{rbv3,nivan\}@cin.ufpe.br.
}

\abstract{%
\highlight{Despite the growing availability of 3D urban datasets, extracting insights remains challenging due to computational bottlenecks and the complexity of interacting with data.
In fact, the intricate geometry of 3D urban environments results in high degrees of occlusion and requires extensive manual viewpoint adjustments that make large-scale exploration inefficient.}
To address this, we propose a view-based approach for 3D data exploration, where a vector field encodes views from the environment. 
\highlight{To support this approach,} we introduce a neural field-based method that constructs an efficient implicit representation of 3D environments.
\highlight{This representation enables both faster \emph{direct queries}, which consist of the computation of view assessment indices, and \emph{inverse queries}, which help avoid occlusion and facilitate the search for views that match desired data patterns.} 
Our approach supports key urban analysis tasks such as visibility assessments, solar exposure evaluation, \highlight{and assessing the visual impact of new developments.}
We validate our method through quantitative experiments, case studies informed by real-world urban challenges, \highlight{and feedback from domain experts.}
\highlight{Results show its effectiveness in finding desirable viewpoints, analyzing building facade visibility, and evaluating views from outdoor spaces.}
Code and data are publicly available at \href{https://urbantk.org/neural-3d}{urbantk.org/neural-3d}.

}


\teaser{
  \centering
  \includegraphics[width=1\linewidth]{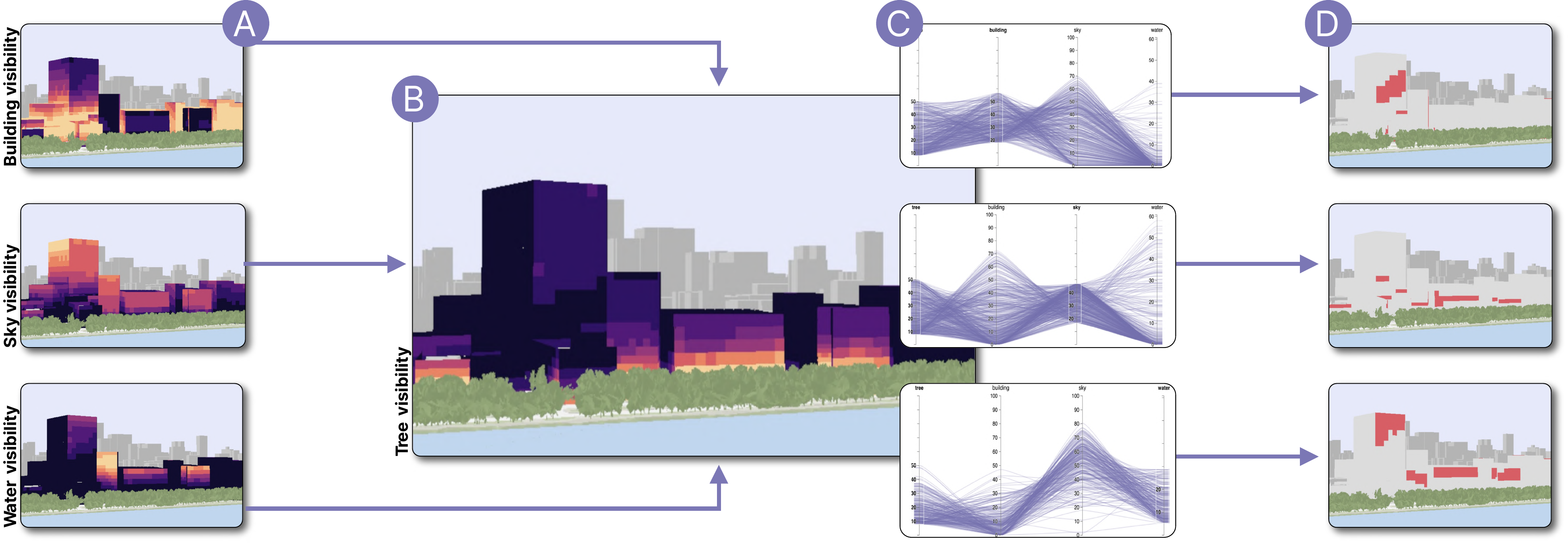}
  \caption{
    Direct queries for the computation of thematic data associated with building facades representing 
    \lower0.2em\hbox{\includegraphics[width=1.1em]{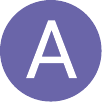}} 
    the visibility towards buildings, sky, and water, and 
    \lower0.2em\hbox{\includegraphics[width=1.1em]{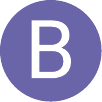}} 
    the visibility towards trees.
    Following 
    the user specified conditions in 
    \lower0.2em\hbox{\includegraphics[width=1.1em]{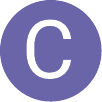}}
    the parallel coordinates we perform 
    \lower0.2em\hbox{\includegraphics[width=1.1em]{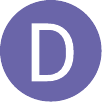}} 
    inverse queries to infer the building facade positions satisfying the joint visibility constraints.
  }
  \label{fig:teaser}
}

\newcommand{\hide}[1]{}

\graphicspath{{figs/}{figures/}{pictures/}{images/}{./}} 

\usepackage{tabu}                      
\usepackage{booktabs}                  
\usepackage{lipsum}                    
\usepackage{mwe}                       

\usepackage{mathptmx}                  

\usepackage{xcolor}

\usepackage{amsmath}
\usepackage{amsfonts}
\usepackage{amssymb}
\usepackage{amsfonts}
\usepackage{amssymb}
\usepackage{algorithm}
\usepackage{algpseudocode}

\begin{document}



\maketitle


\newcommand{\figEvaluationAccuracy}{
\begin{figure}[t]
    \centering
    \includegraphics[width=\linewidth]{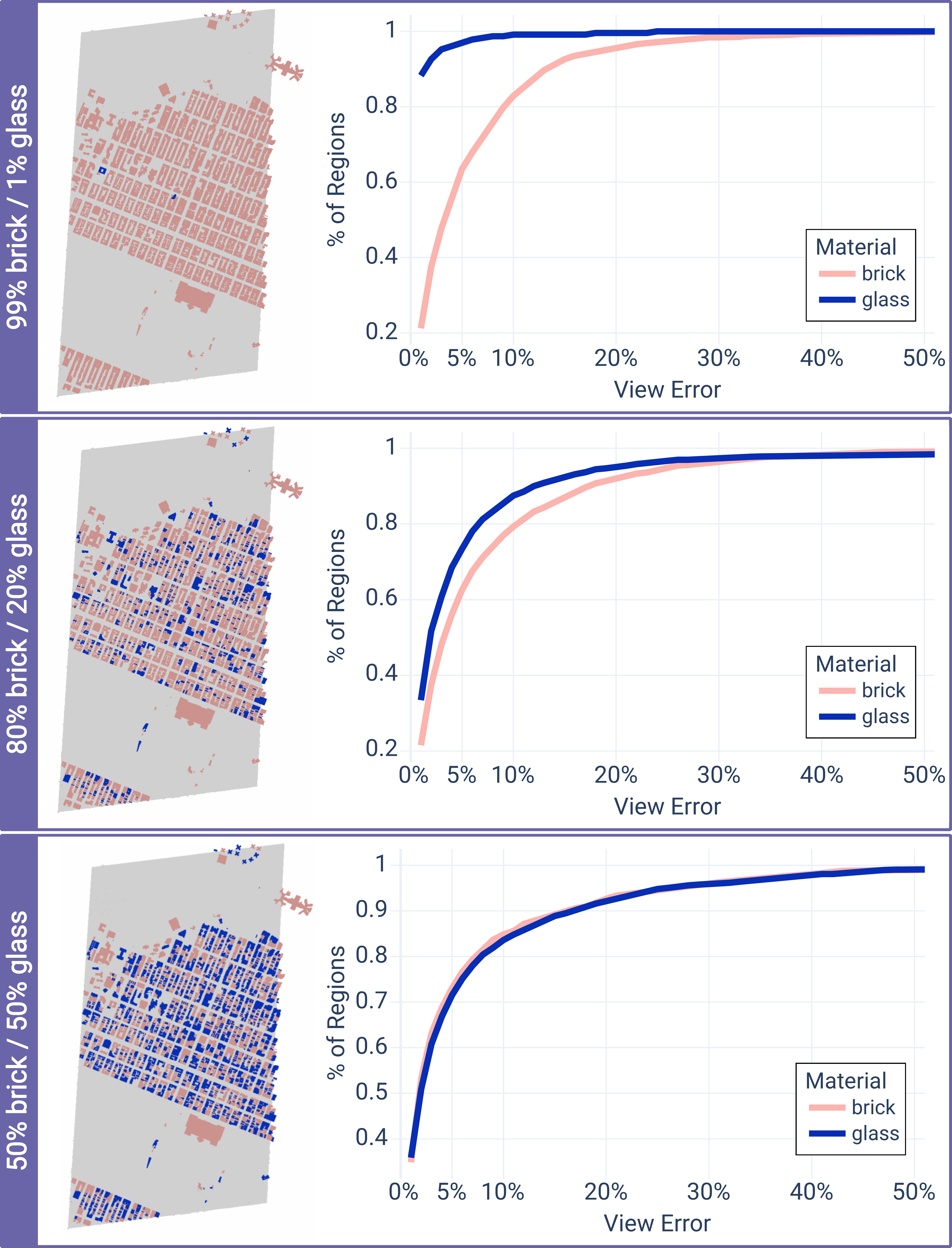}
    \caption{
    Three scenarios for the thematic distribution of building materials.
    On the left side we display aerial views of the scene.
    On the right we show the cumulative view error of our model with respect to the regions of the scene.
    In all three scenarios the model predicts the view thematic composition with less than $10\%$ error for  $\approx 80\%$ of regions in the scene.}
    \label{fig:evaluation_accuracy}
\end{figure}
}


\newcommand{\figEvaluationComparison}{
\begin{figure}[t]
    \centering
    \includegraphics[width=\linewidth]{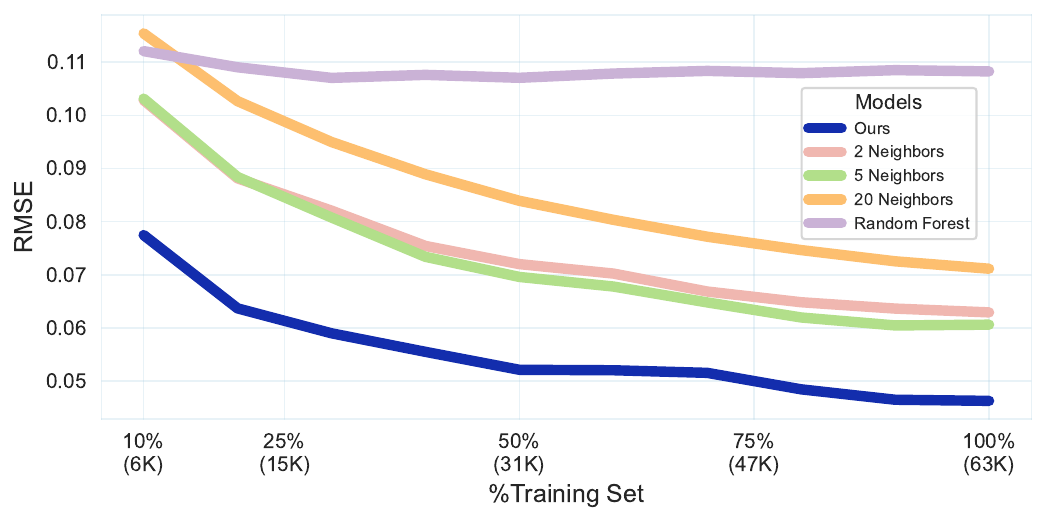}
    \caption{RMSE of our model compared to baseline small-footprint ML models across varying training set sizes.}
    \label{fig:evaluation_comparison}
\end{figure}
}


\newcommand{\figTool}{
\begin{figure*}[t]
    \centering
    \includegraphics[width=0.9\linewidth]{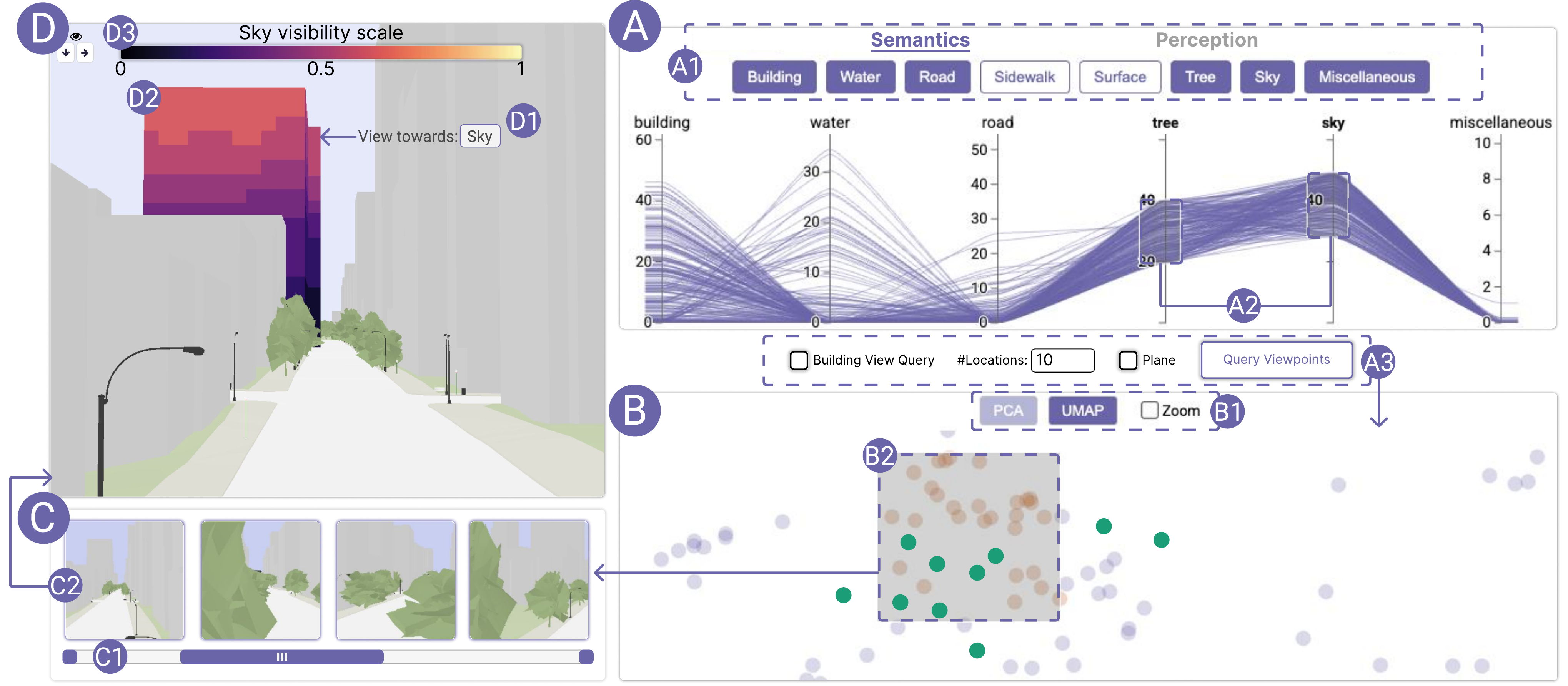}
    \caption{
    The interface includes four main components: \lower0.2em\hbox{\includegraphics[width=1.1em]{figs/icons-pdf/a_svg-tex.pdf}} Control query panel with a PCP for visualizing ground truth visibility distributions. Users can \lower0.2em\hbox{\includegraphics[width=1.1em]{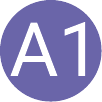}} select distribution type (semantics/perception), \lower0.2em\hbox{\includegraphics[width=1.1em]{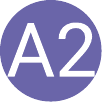}} brush axes to define query constraints, and \lower0.2em\hbox{\includegraphics[width=1.1em]{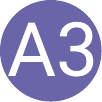}} specify a hyperplane and number of locations to generate new camera views using the model. \lower0.2em\hbox{\includegraphics[width=1.1em]{figs/icons-pdf/b_svg-tex.pdf}} Latent map view shows model predictions (purple) and generated views (green) in latent space; users can \lower0.2em\hbox{\includegraphics[width=1.1em]{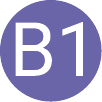}} change projection, zoom/pan, and \lower0.2em\hbox{\includegraphics[width=1.1em]{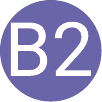}} brush to highlight points. \lower0.2em\hbox{\includegraphics[width=1.1em]{figs/icons-pdf/c_svg-tex.pdf}} Gallery view displays the model-generated camera views associated with brushed green points (from \lower0.2em\hbox{\includegraphics[width=1.1em]{figs/icons-pdf/b2_svg-tex.pdf}}) in \lower0.2em\hbox{\includegraphics[width=1.1em]{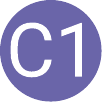}} a scrollable panel, and allows \lower0.2em\hbox{\includegraphics[width=1.1em]{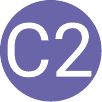}} selecting a view to update \lower0.2em\hbox{\includegraphics[width=1.1em]{figs/icons-pdf/d_svg-tex.pdf}} the 3D map view. The 3D scene supports navigation, with options to \lower0.2em\hbox{\includegraphics[width=1.1em]{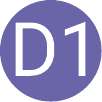}} compute visibility of elements like sky, tree, water from building facades, \lower0.2em\hbox{\includegraphics[width=1.1em]{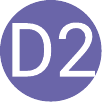}} visualize it on facade tiles, with \lower0.2em\hbox{\includegraphics[width=1.1em]{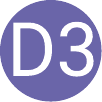}} a reference color legend. 
    }
    \label{fig:tool}
\end{figure*}
}




\newcommand{\figMethodology}{
\begin{figure}[t]
    \centering
      \includegraphics[width=\linewidth]{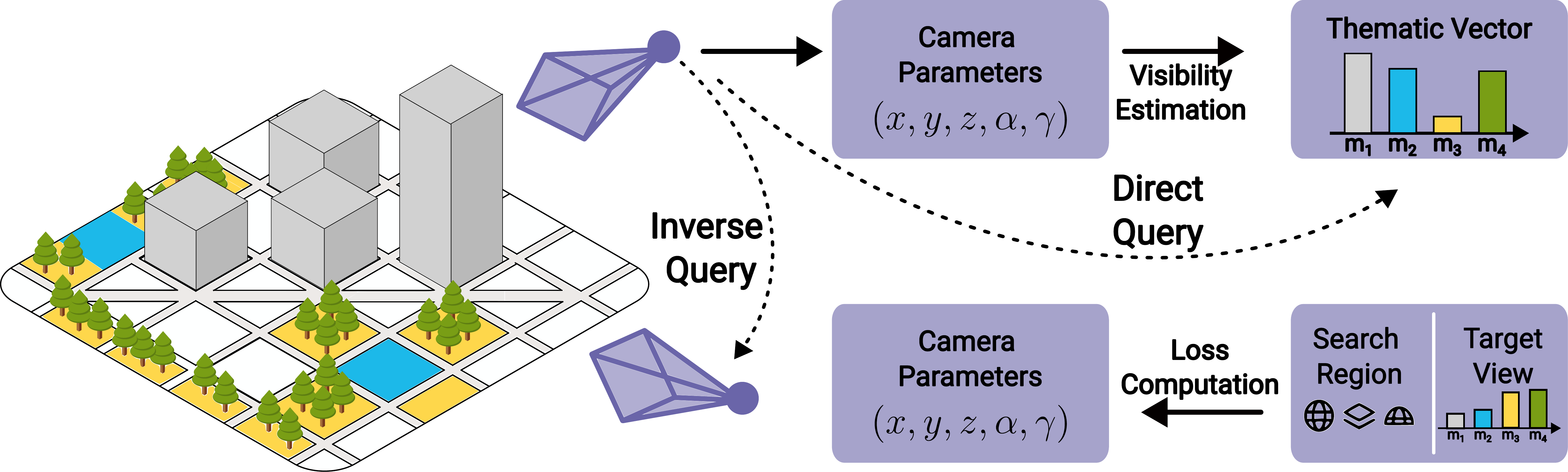}
    \caption{Our proposed visibility-based exploration is enable by two types of queries supported by a machine learning model. 
    Direct queries (top), computing a thematic vector from spatial positions.
    Inverse queries (bottom), searching spatial positions satisfying given target view constraints.
    }
    \label{fig:methodology}
\end{figure}
}






\newcommand{\figUseCaseInverseQueries}{
\begin{figure}[t]
    \centering
      \includegraphics[width=\linewidth]{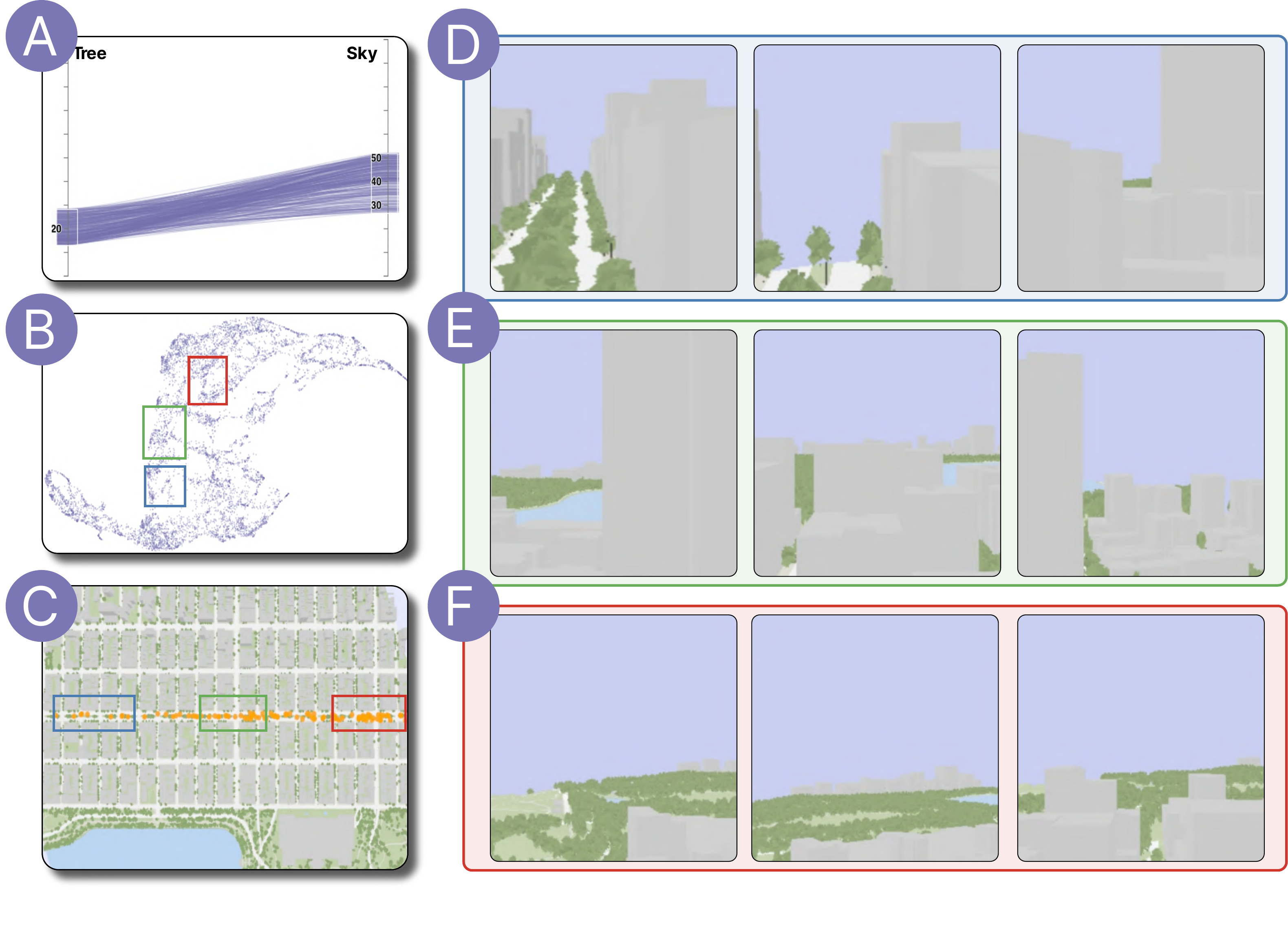}
    \caption{
    Inverse query for views towards 
    \lower0.2em\hbox{\includegraphics[width=1.1em]{figs/icons-pdf/a_svg-tex.pdf}} trees and sky located on 
    \lower0.2em\hbox{\includegraphics[width=1.1em]{figs/icons-pdf/c_svg-tex.pdf}} a street.
     We identify viewpoint clusters using  
     \lower0.2em\hbox{\includegraphics[width=1.1em]{figs/icons-pdf/b_svg-tex.pdf}}
     the latent representation of each location generated by our model.
    In \lower0.2em\hbox{\includegraphics[width=1.1em]{figs/icons-pdf/d_svg-tex.pdf}}, \lower0.2em\hbox{\includegraphics[width=1.1em]{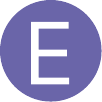}}  and  
\lower0.2em\hbox{\includegraphics[width=1.1em]{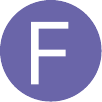}}   
we show a selection of views found through the inverse query. 
    }
    \label{fig:use_case_inverse_queries}
\end{figure}
}

\newcommand{\figUseCaseFromBuildings}{
\begin{figure*}[t]
    \centering
      \includegraphics[width=.9\linewidth]{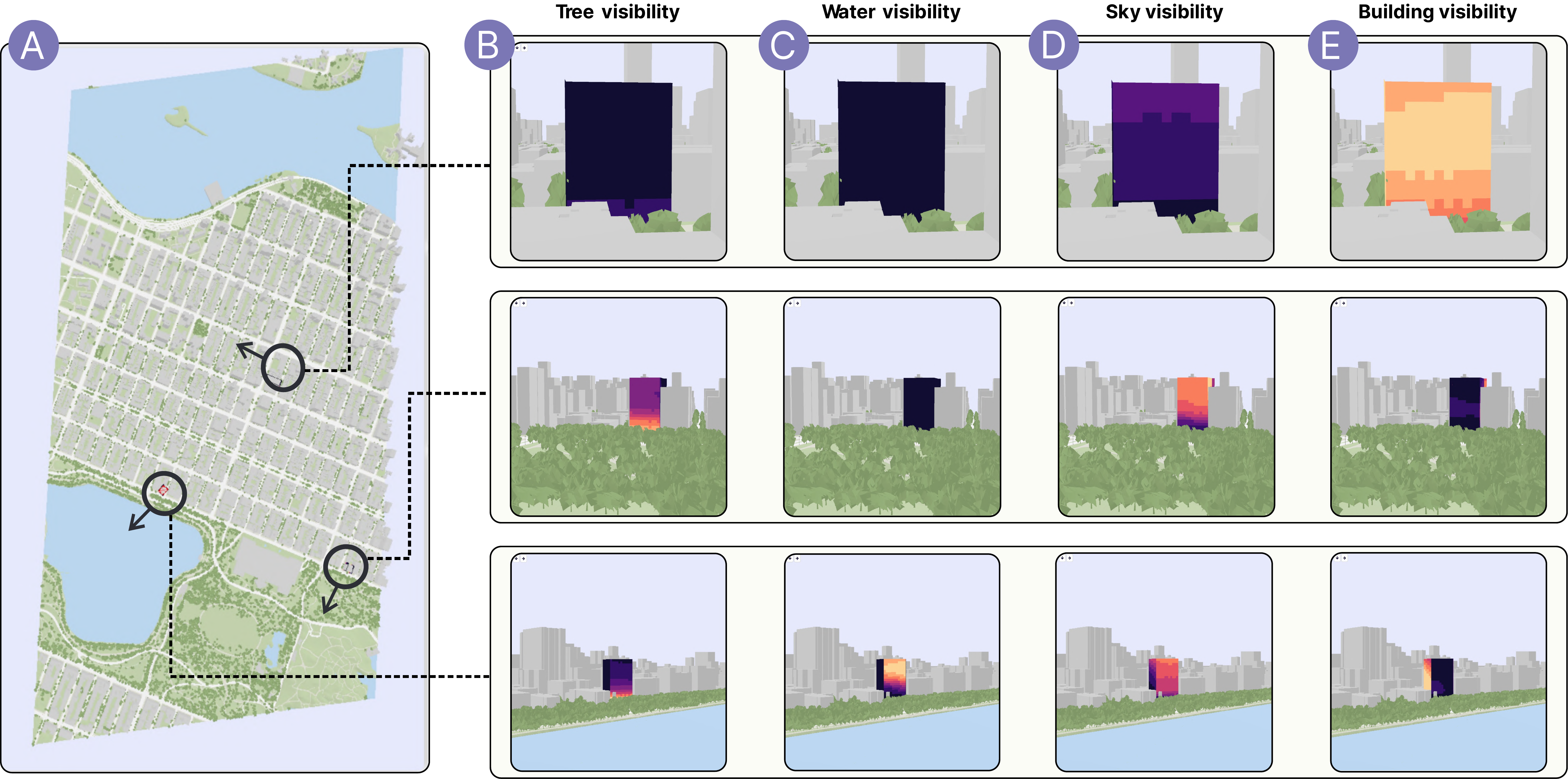}
    \caption{
    Comparative direct queries for the thematic data associated with the facades of three buildings.
    In 
\lower0.2em\hbox{\includegraphics[width=1.1em]{figs/icons-pdf/a_svg-tex.pdf}}
we show an aerial view of the city with the highlighted locations for each building.  
    In columns \lower0.2em\hbox{\includegraphics[width=1.1em]{figs/icons-pdf/b_svg-tex.pdf}}
\lower0.2em\hbox{\includegraphics[width=1.1em]{figs/icons-pdf/c_svg-tex.pdf}},     \lower0.2em\hbox{\includegraphics[width=1.1em]{figs/icons-pdf/d_svg-tex.pdf}},  
and
\lower0.2em\hbox{\includegraphics[width=1.1em]{figs/icons-pdf/e_svg-tex.pdf}}  
we show the thematic data generated for each facade representing the intensity of views towards trees, water, sky and buildings respectively.}
    \label{fig:use_case_from_buildings}
\end{figure*}
}

\newcommand{\tableInterpolationComparison}{
\begin{table}[tb]
\caption{RMSE of our model compared to KNN and Random Forest baselines across varying training set sizes.}
\label{tab:interpolation_comparison}
\scriptsize
\centering
\begin{tabular}{lccccc}
\toprule
\textbf{Model} & \textbf{6k} & \textbf{12k} & \textbf{19k} & \textbf{50k} & \textbf{63k} \\
\midrule
2-Neighbors     & 0.103 & 0.088 & 0.082 & 0.065 & 0.063 \\
5-Neighbors     & 0.103 & 0.088 & 0.081 & 0.062 & 0.061 \\
20-Neighbors    & 0.115 & 0.103 & 0.095 & 0.075 & 0.071 \\
Random Forest   & 0.112 & 0.109 & 0.107 & 0.108 & 0.108 \\
\midrule
\textbf{Ours}   & \textbf{0.078} & \textbf{0.064} & \textbf{0.059} & \textbf{0.048} & \textbf{0.046} \\
\bottomrule
\end{tabular}
\end{table}
}
\section{Introduction}

%
%

\highlight{
The increasing availability of urban datasets with 3D spatial components has created new opportunities for analyzing complex urban phenomena~\cite{miranda_survey_3d_2024}.
Among these possibilities are applications such as visibility evaluation~\cite{yasumoto_use_2011}, view analysis of environmental features~\cite{yu_view-based_2016,virtanen_near_2021,zhang_2023_uncovering}, visibility planning~\cite{ferreira_2015_urbane,ortner_2017_visaware}, view preferences~\cite{kasraian2021evaluating,quintana2024_buildsys_perception}, and building design~\cite{doraiswamy_2015_catalogue,tarkhan2024facade}, which often rely on computing view indices or identifying viewpoints that satisfy specific criteria.
Conventional methods for computing these indices typically rely on ray casting~\cite{miranda_Shadow_2019} or rasterization~\cite{ferreira_2015_urbane}, both of which are computationally expensive and scale poorly to large urban settings.
To mitigate this, many studies either perform offline analyses~\cite{ferreira_2015_urbane,salimi_visual_2023,miranda_survey_3d_2024} or rely on pre-computation to enable interactive exploration~\cite{doraiswamy_2015_catalogue}; however, both approaches ultimately limit true interactivity in visual analysis.
Beyond computation, exploring and interacting with 3D urban datasets presents additional difficulties.
Spatial occlusion and complex camera controls often hinder interpretation: once indices are obtained, they are commonly visualized as thematic overlays on physical surfaces (Fig.~\ref{fig:teaser}(A,B)).
Unlike 2D visualizations with fixed perspectives, 3D exploration requires careful viewpoint selection and manipulation across multiple degrees of freedom.
Without appropriate support, users may struggle to efficiently navigate, reveal patterns, and interpret the data.
Although prior research has extensively explored methods for navigating and visualizing 3D urban environments, much of this work has focused on the representations of cities rather than the interactive analysis of 3D urban data itself~\cite{LEI_challenges_survery_2023}.
For instance, recent studies have investigated viewpoint selection~\cite{boorboor2024submerse}, occlusion-aware rendering~\cite{Alfakhori_occlusion_2022}, and deformation-based de-occlusion and navigation support~\cite{chen_UrbanRama_2022}.
However, these approaches only consider physical data, such as building geometry and infrastructure, while overlooking the integration of thematic attributes.
%
%
As a result, two intertwined challenges arise: (1) the \textbf{efficient computation} of view-related measures that capture the relationships between physical elements (e.g., building geometry, terrain) and thematic attributes (e.g., sunlight access, visibility), and (2) the \textbf{effective exploration} of such measures in 3D environments, where interaction and interpretation are inherently more complex than in 2D settings.

}

%

%

%
%
%


\highlight{To address these challenges, we propose a view-based approach for the interactive exploration of physical and thematic dimensions in this work.
At the core of our method is the modeling of views as a dense multi-dimensional field, where each viewpoint is associated with a semantic attribute vector that encodes relevant properties across large urban environments and provides a unified representation for analyzing spatial patterns.}
%
%
\highlight{Our proposal specifically targets mesh-based urban representations, where surfaces can be directly associated with thematic attributes.}
This representation enables users to assess variations in data distribution, identify patterns, and make comparisons within the urban environment.
%
%
%
\highlight{To efficiently support the operations described above,} we introduce a neural field-based approach, inspired by the well-established neural radiance fields~\cite{mildenhall_2021_nerf}, where a deep neural network learns the vector field. 
\highlight{
In fact, this model enables estimating the vector field value (i.e., semantic view assessment indices) without using additional rendering or ray casting steps.
Furthermore, the differentiable nature of this neural representation allows for the introduction of an algorithm for identifying viewpoints that follow criteria defined by the thematic data, which eliminates} the need for explicit field computation across each viewpoint, leveraging the network’s ability to approximate the field efficiently.
In addition to enabling efficient exploration of 3D urban data, our method can be leveraged for a wide range of applications. 
By encoding physical and thematic attributes in the vector field, the approach supports the analysis of sightlines, occlusion, and visibility across large urban areas.
%

To evaluate our approach, we conducted a set of quantitative experiments to assess its performance and accuracy. Specifically, we measured how quickly views are generated and how accurately they capture the intended spatial relationships within the urban environment.
Beyond these assessments, \highlight{we have implemented a visual interface (see Fig.~\ref{fig:teaser}) and through it we} present three case studies to demonstrate the applicability of our approach in addressing key challenges faced by urban experts.
First, we illustrate how our method can be used to discover novel viewpoints by searching for custom visibility distributions.
Second, we present an efficient building facade visibility analysis with views from buildings towards the outdoors, supporting urban planning and architectural design considerations.
And \highlight{third}, we employ our method to explore views from outdoor spaces towards buildings, an important aspect for urban aesthetics and environmental assessments.
\highlight{Finally, we report on the feedback on our model received in a series of interviews with domain experts.}
Code and data are publicly available at \href{https://urbantk.org/neural-3d}{urbantk.org/neural-3d}.
Our contributions can be summarized as follows:

\begin{itemize}[noitemsep, topsep=0pt, leftmargin=*]
    \item We present a novel approach that models \emph{views} as a vector field encoding both physical and thematic urban attributes.
    \item We propose a neural field–based method that efficiently learns and approximates the vector field.
    \item \highlight{We demonstrate the versatility of our model through experimental evaluation, three practical usage scenarios, and expert feedback received in a series of interviews}.
\end{itemize}

\section{Related Works}

This section reviews previous research related to various aspects of this work. In particular, we review contributions and challenges in visual analytics in 3D urban environments, navigation and occlusion management techniques, and viewpoint estimation.

\subsection{Visual analytics for 3D urban data}

Visual analytics for 3D urban data is a research field motivated by the fact that many phenomena of interest in urban studies are inherently three-dimensional, requiring reasoning over the 3D structure of urban environments.
For this reason, applications in domains like civil engineering, urban planning, architecture, and climate science involve not only data processing of 3D urban data but also using 3D visual representations to properly represent the spatial context of the studied data.
In these scenarios, thematic data layers (e.g., sunlight access, visibility scores) are associated with physical ones (e.g., building geometry, streets, terrain).
As extensively discussed in recent works by Mota et~al.~\cite{mota2023comparison} and Miranda et~al.~\cite{miranda_survey_3d_2024}, unlike the 2D case, there is a lack of visual metaphors designed explicitly for 3D urban environments, as well as a scarcity of evaluation studies that offer grounded design and usage guidelines for such metaphors.
In fact, incorporating the third dimension significantly increases the complexity of designing effective interactive visual representations.
First, features such as perspective distortion and aesthetic visual effects like shading can interfere with critical visual channels commonly used for data representation.
Second, and more relevant to this work, the typical spatial arrangement and density of physical structures in urban environments lead to significant occlusion.
Altogether, these factors contribute to the complexity of data exploration in 3D urban settings.

Elmqvist and Tsigas's taxonomy~\cite{elmqvist2008taxonomy} describes the different modes of occlusion and categorizes de-occlusion methods.
Building on their work, Miranda et al.~\cite{miranda_survey_3d_2024} categorized occlusion handling methods applied to 3D urban analytics into six categories: \emph{bird's view}, \emph{multi-view}, \emph{ghosting}, \emph{deformation}, \emph{slicing}, and \emph{assisted steering}.
Despite the diversity of methods this classification suggests, these rarely find their way into 3D urban analytics solutions. In fact, most previous works rely on the users taking a bird's view through manual camera manipulation to navigate the environment.
%
%
The other categories consist of providing a multi-perspective view (multi-view), 
removing occluding parts via transparency (ghosting), deforming physical data layers (deformation), and slicing volumetric datasets (slicing).
Ghosting techniques are limited due to the difficulty in handling multiple layers of occluders~\cite{deng_interactive_2016}.
Deng et al.~\cite{deng_interactive_2016} proposed a hybrid approach that applies various deocclusion operators to a given viewpoint. The operators include bird’s-eye view adjustments (increasing the camera height) and deformation techniques such as road shifting, building scaling, and building displacement.

On the other end of the spectrum, assisted steering techniques were found to be the least prevalent in 3D urban visualization systems.
These techniques automatically compute optimal viewpoints according to some criteria.
The solution proposed in our paper falls in this category.
As Miranda et al.~\cite{miranda_survey_3d_2024} surveyed, the assisted steering methods applied to 3D urban environments have two main limitations. First, they only consider physical layers in their optimality criteria.  
Second, they can be computationally expensive, particularly when relying on non-differentiable viewpoint optimality criteria~\cite{boorboor2024submerse}.
Our proposed method consists of a deep learning-based approach that enables fast and flexible selection of viewpoints based on thematic data layers.

\subsection{Visibility exploration in 3D urban environments}
Many essential use cases are targeted by 3D urban visual analytics studies~\cite{miranda_survey_3d_2024}. 
Among them, visibility analysis is very relevant to our work.
%
Domain works focusing on this area tackle problems like view access equity~\cite{yasumoto_use_2011}, visibility to green spaces~\cite{yu_view-based_2016,virtanen_near_2021}, blue spaces~\cite{zhang_2023_uncovering}, and landmarks~\cite{salimi_visual_2023}, views of high rises~\cite{li_room_2022}, and office rents~\cite{turan_development_2021}.
%
%
In addition, some works in the visualization community focused on designing visual analytics systems for visibility analysis. 
For example, the works by Ortner et al.~\cite{ortner_2017_visaware} and Ferreira et al.~\cite{ferreira_2015_urbane} presented design studies to support visibility-aware urban planning and view impact analyses, respectively.
%
%
These works relied on a rasterization strategy to compute visibility scores. 
However, since visibility computation is expensive, these could impact the system's interactivity when operations requiring many scene renderings are executed.
For this reason, different strategies were proposed.
For example, Doraiswamy et al.~\cite{doraiswamy_2015_catalogue} proposed a computational topology-based framework for building design.
This work relied on a pre-computation-based strategy to support interactive visualization and real-time building design optimization.
The results of the pre-computation were stored in a regular grid data structure.
Closer to our work, Li et al.~\cite{li_room_2022} used a deep learning-based model to estimate visibility-based scores to assess the quality of the views from the windows in building facades. 
While close in intent, they employ a different modeling strategy than ours.
In fact, they perform a pixel-level segmentation step followed by a regression phase to compute the visibility scores.
Our model reduces this computation by directly computing the semantic aggregation from the views (without the segmentation step).
Finally, unlike previous works, our model is used not only to accelerate visibility score computation but also to enable semantic search over viewpoints.



\subsection{Viewpoint selection techniques}

Viewpoint selection has been a widely studied problem in different fields.
When view targets are clearly defined, direct methods can be used to place the camera to maximize views towards this target.
These methods consist of heuristics that compute camera parameters from the target's spatial position and extent.
For example, Li et al.~\cite{li_2017_viewpoint} introduced a 3D visualization technique for trajectory attribute data, combining it with a geometric viewpoint selection strategy to either focus on specific trajectory segments or provide an overview with a broader viewing area.
In more complex scenarios, \emph{optimization-based methods} are the most common.
This approach involves defining a quality measure for a given viewpoint and using optimization algorithms to select the best viewpoint based on that measure.
%
%
Bonaventura et al.~\cite{bonaventura2018survey} presented a survey of viewpoint selection quality measures for polygonal models, primarily targeting computer graphics applications.
%
The measures analyzed focus on assessing geometric properties (e.g., area, silhouette, depth, surface curvature) of a polygonal mesh as perceived from a given viewpoint.
%
%

Few works have addressed viewpoint selection in 3D urban environments.
Li et al.~\cite{li_seevis_2020} followed the direct approach to generate customized and narrative animations in an evacuation planning context. 
Neuville et al.~\cite{neuville_2019_3dviewpoint} proposed a technique that uses visibility to user-defined points of interest as a quality measure. 
Viewpoints are sampled in a hemisphere around the points of interest, and the quality measure is computed for all these points to assess the best one.
Zhang et al.~\cite{zhang_urbanvr_2021} and Boorboor et al.~\cite{boorboor2024submerse}  built an immersive analytics system for site development and flood simulation analysis, respectively. They followed the optimization-based approach to select viewpoints towards a set of user-defined points of interest. This was done by optimizing an energy function that considers geometrical visibility and spatial location aspects. 
In all of these techniques, the viewpoint selection criteria only considered physical elements of the city and also focused on specific geometric elements (e.g., buildings) as targets.
Our technique differs from these in two key ways. First, it is designed with a different goal in mind. More precisely, we are interested in assisting the exploration of 
\emph{semantic view indices},
i.e., views considering elements of thematic data layers. 
In our approach, the user specifies semantic elements of what they desire to visualize instead of physical targets.
Second, we avoid expensive visibility computations by using a neural network approximating the visibility given a viewpoint. 
As detailed in Section~\ref{sec:technique}, our model is trained on a relatively small set of viewpoints and approximates the desired visibility function well.

Other works also approached the view selection problem via machine learning methods.
These were proposed mainly in the computer vision community~\cite{genova_2017_learning}.
%
%
Sun et al.~\cite{Sun_2021_CVPR} proposed optimizing multi-camera placement for maximum visibility coverage.
Their method approximates visibility coverage using a neural network and reformulates the multi-camera placement problem into a continuous optimization task.
While their contributions are similar to ours, we adopt a different modeling strategy inspired by neural radiance fields.
To the best of our knowledge, our work represents the first machine learning-based approach applied specifically to 3D urban visual analytics.
Moreover, our highly flexible model supports evaluating a wide range of previously proposed metrics and enabling diverse analysis tasks.




\section{Background \& Challenges}

In this paper, we investigate a new approach for two intertwined problems in 3D urban data exploration: view estimation and visibility analysis.
3D urban data is often derived from sources such as LiDAR scans, simulations, and modeling, and it provides a representation of the urban environment, capturing both physical and thematic attributes relevant to buildings and infrastructure.
These datasets are widely used in urban planning~\cite{RANZINGER1997159}, environmental assessments~\cite{Fol_environmental_2024}, and infrastructure planning~\cite{MIRZAEI2022101501}, but their complexity makes extracting meaningful insights challenging.
Our recent survey summarized a series of domains, tasks, and data types associated with 3D urban analysis, highlighting key challenges in effectively leveraging these datasets~\cite{miranda_survey_3d_2024}.
One fundamental aspect is the interplay between physical and thematic data. Physical data, such as the geometry of buildings and infrastructure, define the structural environment, while thematic data, often derived from simulations, encodes additional attributes like solar exposure, air quality, or noise levels.
This interplay leads to occlusion problems, as thematic overlays are often projected onto physical layers (e.g., building surfaces), making them difficult to analyze when obstructed from certain viewpoints. However, both physical and thematic data are essential for analyses~\cite{miranda_survey_3d_2024}, meaning that effective exploration must account for their spatial relationships and visibility constraints.

Visualization and visual analytics systems that leverage such data often rely on manual camera navigation to resolve these issues~\cite{ferreira_2015_urbane,doraiswamy_2015_catalogue,willenborg_applications_2017,miranda_Shadow_2019}, which means that the user must adjust viewpoints iteratively to reveal occluded thematic information.
This is cumbersome and time-consuming.
%
Unlike 2D visualizations, 3D environments demand constant interaction (panning, zooming, rotating) to explore structures of the physical and thematic overlays, with users often resorting to trial-and-error navigation to access relevant data.
In a large-scale urban environment, such an approach is particularly inefficient, as relevant data may be scattered across multiple locations.
View estimation can address this problem by determining optimal viewpoints, supporting a more effective exploratory process while reducing reliance on exhaustive manual adjustments.
Fast, automated view estimation can enhance interactive analysis by guiding users toward informative perspectives.

%
Parallel to this, visibility analysis is a domain-specific problem that focuses on analyzing the visibility of urban structures and natural features from a set of viewpoints. 
Thus, visibility analysis requires the computation of new data that quantifies which elements of the built environment are visible from given locations.
This process has applications in solar exposure assessment~\cite{florio_2021_solarenergy}, urban design evaluations~\cite{perry_visibility_urban_evaluation_2007}, and sightline analysis~\cite{Chen2017ImmersiveUA}.
However, traditional visibility analysis techniques, such as exhaustive ray casting, are computationally expensive and scale poorly to large urban settings.
This process can also benefit from an effective view estimation computation.

Our approach integrates fast view estimation as a means to enhance both data exploration and visibility analysis.
At its core, we create an implicit representation of the 3D environment. Such representation contains, for a given position, an estimate of the data distribution seen from that position, incorporating both physical and thematic attributes.
By employing an implicit approach, we are able to perform fast computation of requested viewpoints to support interactive data exploration.

\figMethodology

\section{View-Based 3D Urban Data Exploration}
\label{sec:technique}

Motivated by the aforementioned challenges and the growing availability of 3D urban data, we propose an alternative method for exploring these datasets, shifting away from traditional navigation-based exploration towards a view-driven, queryable framework.
To support this view-driven framework, at the core of our work is the need to \emph{estimate views}, i.e., compute a distribution capturing both physical and thematic data visible from a given viewpoint. Such estimation of views will support an exploration of the 3D urban environment without the need for manual navigation.
To achieve this, our method builds a neural implicit representation of the 3D environment, where per-view data distributions are encoded as a continuous vector-valued function.
This representation supports: (1) \textbf{direct queries}, where users retrieve metrics for any viewpoint without requiring scene navigation or expensive ray tracing, and (2) \textbf{inverse queries}, where users specify a desired data pattern and retrieve viewpoints that satisfy these criteria.
%
%
Our approach adopts a methodology inspired by the well-established neural radiance fields~\cite{mildenhall_2021_nerf}.
However, instead of predicting radiance and opacity along a camera ray to synthesize an image per viewpoint, our model learns to approximate the distribution of data metrics per viewpoint, capturing both physical visibility and thematic attributes of the environment.
In other words, instead of reconstructing an image, we learn a continuous function that maps spatial coordinates and viewing directions to data distributions. This allows us to retrieve information about a scene rather than reconstruct its appearance (see Fig.~\ref{fig:methodology}).
Next, we detail our approach, first covering our vector field representation (Section~\ref{sec:function}) followed by our model to learn an implicit representation of this vector field (Section~\ref{sec:model}).

\subsection{Vector-valued function representation}
\label{sec:function}

We define a viewpoint as a tuple $(x,y,z,\alpha,\gamma) \in \mathbb{R}^5$, where $(x,y,z)$ denotes the 3D spatial position in Cartesian coordinates and $(\alpha, \gamma)$ is a viewing direction represented by Euler angles (we ignore roll).
The set of all viewpoints is denoted by $\mathcal{V}$.
We assume that the physical layers describing the urban environment are represented as polygonal meshes.
These include both nature elements (e.g., trees, bodies of water) and man-made elements (e.g., roads, buildings).
Furthermore, we assume that we are given a thematic data layer represented by a vector field $T$ defined on the vertices of the mesh representing the elements of our physical layer.
This vector field sets the semantic context for our analysis and depends on the specific application. 
For example, in a sunlight access study, $T$ might represent the amount of direct sunlight received on a building surface. 
In a visibility study, we could classify the different physical elements into categories (e.g, buildings, roads, trees, and water) and assign via $T$ an integer value for each category.

To represent the 3D urban environment, we define a vector-valued function representation that encodes both spatial structure and view-dependent thematic data.
%
We call this function $S: \mathcal{V} \to \mathbb{R}^k$, where $S(x,y,z,\alpha,\gamma) \in \mathbb{R}^k$ produces a vector $\mathbf{m} = (m_1,m_2,...,m_k)$ that encodes $k$ scalar values describing what can be observed from that viewpoint.
%
%
%
%
To define the value of $\mathbf{m}$ for a given viewpoint, we first rasterize the 3D scene, producing a 2D image where each pixel encodes the physical elements and the thematic vector of view indices $T$ from the surfaces visible from that viewpoint. 
The image will provide a visual summary of the environment as perceived from the given viewpoint.
After rendering the rasterized image, we extract a distribution of view indices by aggregating pixels into bins.
In fact, each component $m_i$ of $\mathbf{m}$ represents the aggregated count of pixels falling into a specific range within the view indices.
Formally, we denote by  $I\vcentcolon= I_{(x,y,z,\alpha,\gamma)}$ the rasterized image for a given viewpoint, and $(u,v)$ represents the pixel coordinates, $I(u,v) \in \mathbb{R}$ is a thematic attribute observed in that pixel.
The distribution of view indices will be given by $m_i = \sum_{(u,v)} \mathbbm{1}_{B_i}(I(u,v))$, where $m_i$ is the aggregated value for a bin $B_i$, considering a set of bins $\mathbf{B}$, 
and $\mathbbm{1}_{B_i}$ is the indicator function of the bin $B_i$. 
The sum accumulates the thematic data within each bin across all pixels.
Finally, the counts in the bins are normalized so the sum of the values adds up to 1.

For example, in a sunlight access study, the thematic value $T$ might represent the amount of direct sunlight received on a building surface. 
In this scenario, we could define three bins ($B_1$ for 0-2 hours of sunlight, $B_2$ for 2-5 hours, and $B_3$ for five or more hours) and the resulting feature vector $\mathbf{m}=(m_1, m_2, m_3)$ will indicate the proportion of visible surfaces that belong to each bin.
Such representation naturally encodes contributions to a given thematic attribute, taking into account visibility constraints.
Moreover, by taking into account the distribution of thematic values across viewpoints, this representation allows us to retrieve camera positions that meet certain filtering constraints.

Since $S$ encodes distributions for a large number of positions and viewing directions in a 3D environment, it results in a high-dimensional and large-scale representation that might span large areas with millions of potential viewpoints. 
Therefore, storing and processing this function explicitly becomes impractical.
To address this computational challenge, we propose a compact and efficient way to store and query $S$ without explicitly maintaining the original data.
Next, we detail how we use an implicit neural representation, which learns a function approximation of $S$, reducing storage usage while enabling efficient retrieval of thematic distributions.

\subsection{Neural field-based implicit representation}
\label{sec:model}


To represent the vector-valued function $S$ without explicitly storing all values, we adopt an implicit neural representation inspired by neural radiance fields.
This representation provides a compressive representation of the visibility function $S$  with a much smaller memory footprint than we would need to store the raw data~\cite{lu_2021_compressive}.
However, unlike traditional neural radiance fields, which model radiance and opacity to synthesize images, we approximate the distribution of thematic values visible from any given viewpoint in the 3D environment.
In other words, instead of storing discrete values, we train a neural network $F_\Theta: \mathcal{V} \to \mathbb{R}^k$ that will approximate the original vector field.

\subsubsection{Model architecture}
$F_\Theta$ is implemented as a multi-layer perceptron that takes as an input the spatial position and viewing direction and outputs the predicted distribution of view indices, i.e., $\mathbf{m} = F_\Theta(x,y,z,\alpha, \gamma)$. 

To start, each coordinate and viewing direction is embedded using a high-frequency function, which was previously used in the neural radiance fields methodology. 
This type of input embedding was shown to improve the performance of neural networks when learning high-variation data.
Specifically, we project the input using 
$\rho:\mathbb{R} \to \mathbb{R}^{10}, $ 
$\rho(t) = (\sin(2^0\pi t), \cos(2^0\pi t),\dots, \sin(2^9\pi t), \cos(2^9\pi t))$. 
The input obtained for the 3D coordinates $(\rho(x),\rho(y),\rho(z))$ is passed through 10 fully-connected layers, each with 256 nodes and with ReLU activations.
We included a skip connection by concatenating the first layer's inputs with the sixth layer's input.
We used a 128-node layer with a hyperbolic tangent activation function as a final layer. 
This takes as input the direction embeddings $\rho(\gamma),\rho(\theta)$ and the output of the 10-layer network
to return the thematic distribution $\mathbf{m}$.


\subsubsection{Generalized model}
\label{sec:generalized_model}


We generalize the model described above by introducing two additional modules: input regularization and output customization.
First, we show how the input can be restricted to a regular subspace. 
As our input domain is $\mathbb{R}^3$ and our scene is bounded, we can consider that it actually resembles an irregular 3D cube.
Learning subspaces is an extension of learning the values of the function $S$ in the whole domain.
To exemplify, we consider bounded 2D hyperplanes as parallelograms contained in the cube domain. 
We define a parallelogram by its origin $p$, its normalized side direction vectors $(v_1, v_2)$, and lengths $(l, L)$.
These constants induce a parametrization for every point $(x,y,z)$ in the given parallelogram, by finding an $a,b \in [0,1]$ such that $(x,y,z) = \zeta (a, b)$: 

$$ \zeta (a, b) = p + a l v_1 + b L v_2.$$

Following the same steps, the parametrization function $\zeta$ can be replaced by any other parameterizable spatial region in the 3D cuboid. 
In our experiments, we tested spheres and hemispheres, for which $\zeta$ becomes $ \zeta (a, b) = c + (r \cos(a) \sin(b),  r \sin(a)  \cos(b), r \cos(b))$, where $a \in [0,\pi]$ (or $a \in [0,\pi/2]$ for a hemisphere) and $b \in [0,2\pi]$. 
In this context, $c$ and $r$ are constants, representing the sphere's center and radius. 

Second, we introduce an output customization module. 
Our definition for thematic data is based on a fixed representation of the 3D urban environment.
Nonetheless, in urban settings, there is sometimes a need for more abstract metrics to assess the visual experience \cite{ma_2021_perception}. 
For example, the walkability experience can be described as the ratio between pavement and the overall road.
We allow the definition of custom perception thematic metrics as algebraic expressions of the thematic vector $\omega: \mathbb{R}^k \to \mathbb{R}$.
The walkability thematic metric can thus be defined as $m_w =  \omega_w(\mathbf{m}) =  m_{sidewalk} / (m_{sidewalk} + m_{road})$.

Considering the input regularization and output customization modules, the formulation of the general model becomes $F_\Theta = \omega \circ F_\Theta' \circ \rho \circ \zeta$, where $F_\Theta'$ is learnable, $\rho$ is fixed, whilst $\omega$, $\zeta$ are fixed but adjustable by the user. 
The generalized output is a customized thematic vector $\mathbf{m} = F_\Theta (a, b) = \omega \circ F_\Theta' \circ \rho \circ \zeta (a,b) $.
Notably, all functions that compose $F_\Theta$ are differentiable. 
This allows us to train $F_\Theta'$ using standard backpropagation to learn a mapping from viewpoints to the original thematic data.
Importantly, the customization modules $\zeta$ and $\omega$ can be applied at test time without requiring additional training.
Given its differentiable nature, we will refer to $F_\Theta$ as a neural network for the remainder of this paper.


\subsubsection{Training the model}
To train $F_\Theta$, we use a dataset of pairs of viewpoints and their corresponding thematic distributions.
The ground-truth dataset consists of pairs $\{(x_i, y_i, z_i, \alpha_i, \gamma_i),  \mathbf{m}_i^{\text{gt}} \}_{i=1}^{N}$, where $\mathbf{m}_i^{\text{gt}}$ is the vector of thematic value view indices observed from the rasterization of that viewpoint.
The ground-truth data will depend on the specific problem and the thematic attributes of interest.
For example, in a sunlight access analysis, ground truth may be obtained from sunlight simulation data on the surface of buildings. The data may be evaluated by occlusion from different viewpoints for visibility analysis.
This also means that the sampling strategy will depend on the problem. For example, in a sunlight access analysis, viewpoints may be sampled from public open spaces (e.g., parks, streets), while for visibility analysis, viewpoints may be sampled along building surfaces.
We will detail specific cases in Section~\ref{sec:cases}.
The model is trained using the loss function: 
\[
\mathcal{L} = \frac{1}{N}\sum_{i=1}^{N} \left\| F_\Theta(x_i, y_i, z_i, \alpha_i, \gamma_i) - \mathbf{m}_i^{\text{gt}}\right\|_2^2
\]


\subsubsection{Querying the model}
Once trained, $F_\Theta$ provides an efficient way to retrieve thematic distributions for a given viewpoint or find viewpoints satisfying specific conditions.
We name these as direct queries (i.e., forward retrieval of thematic data) and inverse queries (i.e., finding desired viewpoints), as shown in Fig.~\ref{fig:methodology}.


\noindent \textbf{Direct queries.}
%
A direct query will retrieve the thematic distribution $\mathbf{m}$ for a given spatial position and viewing direction, i.e., $\mathbf{m} = F_\Theta(x, y, z, \alpha, \gamma)$.
This is a single forward pass through the network. This is particularly efficient given that the model does not require explicit storage of the vector field; it implicitly represents the field as a function.
This allows users to quickly extract information about what is visible from a given viewpoint.

\begin{algorithm}[t]
\caption{Gradient-Based Optimization for Inverse Query}
\label{alg}
\begin{algorithmic}[1]
\State \textbf{Input:} Initialized viewpoint $(x_0, y_0, z_0, \alpha_0, \gamma_0)$, learning rate $\eta$, target thematic vector of view indices $\mathbf{m}^*$, tolerance $\epsilon$, maximum number of iterations $N$
\State \textbf{Output:} Viewpoint $(x^*, y^*, z^*, \alpha^*, \gamma^*)$ minimizing $\|\mathbf{m} - \mathbf{m}^*\|^2_2$

\setlength{\abovedisplayskip}{0pt} \setlength{\belowdisplayskip}{0pt}
\While{$\|\mathbf{m} - \mathbf{m}^*\|^2_2 > \epsilon$ and iterations < $N$}
    \State Compute gradient of loss function: $\nabla_{x, y, z, \alpha, \gamma} \mathcal{L}$
    \State Update viewpoint using gradient descent:
    \[
    (x^*, y^*, z^*, \alpha^*, \gamma^*) \gets (x, y, z, \alpha, \gamma) - \eta \cdot \nabla_{x, y, z, \alpha, \gamma} \mathcal{L}
    \]
    \State Evaluate new thematic vector:
    \[
    \mathbf{m} = F_\Theta(x, y, z, \alpha, \gamma)
    \]
\EndWhile

\State \textbf{Return} $(x^*, y^*, z^*, \alpha^*, \gamma^*)$
\end{algorithmic}
\end{algorithm}

\noindent \textbf{Inverse queries.}
An inverse query will search for inputs ($x, y, z, \alpha, \gamma$) that minimize the difference between the inferred thematic vector $\mathbf{m}$ and the desired vector of view indices $\mathbf{m}^*$. 
This process involves solving:
\[
(x^*, y^*, z^*, \alpha^*, \gamma^*) = \arg\min_{(x, y, z, \alpha, \gamma)} \|\mathbf{m} - \mathbf{m}^*\|_2^2
\]
We explore two alternatives for performing inverse queries.
First, by choosing a large finite subset of viewpoints out of the total space $\mathcal{V'} \subset \mathcal{V}$.
The thematic distribution for all the viewpoints can be computed efficiently by adjusting the neural network input batch size accordingly.
Out of the computed vectors, the ones closest to the desired vector $\mathbf{m}^*$ can be returned as approximate solutions.
This approach relies on an informed choice for the viewpoint subset $\mathcal{V'}$.

The second approach for these queries is inspired by the process of inverting neural radiance fields~\cite{yen_2021_inerf}.
Given that $F_\Theta$ is a neural network, its output $\mathbf{m}$ is differentiable with respect to its input $(x, y, z, \alpha, \gamma)$.
We then use gradient-based optimization to iteratively adjust the viewpoint so that the inferred  $\mathbf{m}$ becomes as close as possible to the desired target vector $\mathbf{m}^*$.
As a stopping criterion, we use the norm between the inferred thematic vector $\mathbf{m}$ and the desired $\mathbf{m}^*$, finishing the search when $\|\mathbf{m} - \mathbf{m}^*\|_2^2 \leq \epsilon$, where $\epsilon$ is a predefined threshold.
Furthermore, we added a maximum number of iterations parameter, $N$, to allow the user to have better control of the total running time of the algorithm.
To explore multiple candidate viewpoints that satisfy the desired thematic conditions, we initialize the optimization process from a set of random locations sampled across the spatial extent of the city.
This process is detailed in Algorithm~\ref{alg}.




\figTool

\section{Applications in Urban Analyses}
\label{sec:cases}

We now illustrate how our model can support view-based exploration of a 3D urban environment. 
To do so, we use two scenarios to motivate the need for two different tasks.

The first consists of performing visibility analysis on building surfaces (studying \emph{what can be seen from buildings}). This scenario will illustrate the use of \text{direct queries} to build visual summaries. 
The second scenario corresponds to the use of \textbf{inverse queries} to find points that have a particular view pattern, given the view distribution of thematic values.

\subsection{Direct queries for visibility analysis}

Visibility analysis can be performed from different perspectives and at different scales.
For example, an architect might want to analyze what can be seen from viewpoints on the surface of many buildings in a neighborhood.
The surface of each building is usually segmented into a series of patches that define the resolution for the analysis.
This scenario presents two key challenges for interactive visibility data analysis.
First, computing visibility-based indices that describe what can be seen from each building surface patch is computationally expensive.
As a result, a common approach is to precompute these indices before analysis, using a fixed resolution and a predefined set of buildings.
Second, when the number of buildings and the resolution increase, the memory footprint of the resulting visibility data grows significantly.
In such cases, memory usage can become difficult to manage, or even prohibitive, limiting both the efficiency and scalability of the analysis.

The model described in Section~\ref{sec:technique} can be seen as a solution to these challenges. 
In fact, our model provides a compressive representation of the visibility function with a much smaller footprint than we would need to store the raw data~\cite{lu_2021_compressive}.
Furthermore, we can use the model's direct query at arbitrary locations/resolutions within the 
domain of the training data. 
Finally, the computation of the direct query is highly parallelizable, and its computation time is independent of the underlying geometry that represents the physical layers of the city.

%
This makes our model suitable for interactively building a range of customizable visual summaries. 
For example, we can build heatmap layers to display visibility indices on the surface of buildings. 
For this, the user can specify a set of target buildings. 
%
Given this, we use a customizable resolution parameter to build rectangular patches on the surface of each target building, as shown in Fig.~\ref{fig:tool}(D2).
%
We can use a set of direct queries for each patch to summarize the patch visibility index. 
For this, we randomly sample a number of viewpoints $P = \{(x_i,y_i,z_i)\}$ within the patch and use the normal vector of the patch to specify the corresponding Euler angles $(\alpha,\gamma)$.
The thematic value for each patch is computed as $\mathbf{m} = \frac{1}{P}\sum_{i=1}^P F_\Theta(x_i, y_i, z_i, \alpha, \gamma)$. 
%
%

\subsection{Inverse views for pattern-based navigation/exploration}

Planning constructions to take advantage of available views is a key concern in architecture~\cite{ortner_2017_visaware}.
In dense urban environments, views are a scarce and valuable resource.
As a result, architects often need to assess which views are accessible from specific areas within the city.
Moreover, each building project comes with its own set of constraints, requiring architects to adapt their search for suitable views based on the project phase and clients' (potentially evolving) preferences.
In this context, manually evaluating and exploring viewpoint options can become time-consuming or even infeasible.

Our model can be applied as an automated tool for searching views with desired characteristics in this context.
To support this, we offer users three different strategies for conducting such searches.
The first approach is designed to broadly explore views across the city using unconstrained gradient-based inverse queries.
For instance, a user can search for viewpoints that offer a balanced view between buildings and water by specifying an appropriate view index query vector $\mathbf{m} = (m_b, m_w) = (0.33, 0.33)$. 
The model optimizes a set of randomly initialized starting locations to approach the specified $\mathbf{m}$, without imposing any location constraints.
As a result, the identified viewpoints may lie in parks, sidewalks, building facades, or other arbitrary locations throughout the city.

The second approach enables users to specify spatial preferences for exploration and desired view composition $\mathbf{m}$. 
This type of inverse query is useful for identifying visual highlights along a specific street or within a particular area of the city.
To define a target area, the user specifies a plane $\zeta(\cdot, \cdot)$  positioned over the region of interest.
Gradient-based optimization is then performed with respect to the plane parameters $(a, b)$, allowing the model to return locations on the plane -- given by $(x,y,z) = \zeta(a,b)$, where $F_\Theta(x,y,z)$ closely matches the target $\mathbf{m}$.

%
The third approach for performing inverse queries is tailored to well-structured surfaces that meet two key criteria.
First, they have a fixed orientation, meaning the viewing direction is predetermined. Second, they can be divided into tiles or patches with relatively low view variation.
A common example of such surfaces in urban environments is building facades. For each facade, we assume a fixed viewing direction aligned with its surface normal, and we partition the facade into rectangular patches.
By computing visibility scores for each patch, we can identify and filter those that meet a given view constraint.
This method provides an effective way to perform inverse queries for locating desired viewpoints on building facades.

\subsection{Visual interface}
We have implemented a visual interface that leverages our model to support view-based exploration of 3D urban datasets.
The interface, shown in Fig.~\ref{fig:tool}, is comprised of four components.

\noindent \textbf{Control query panel.} 
The panel (Fig.~\ref{fig:tool}(A)) shows the ground truth visibility distribution using a parallel coordinate plot (PCP) with distribution displayed in terms of view indices (e.g., buildings, sky), or perception (e.g., greenness, walkability) (Fig.~\ref{fig:tool}(A1)). 
Users can control which variables are displayed in the PCP by using the toggle buttons positioned above the plot (Fig.~\ref{fig:tool}(A)). The PCP is also used for defining query constraints -- such as specifying tree visibility between 20–40\% and sky visibility between 30–50\% through brushing (Fig.~\ref{fig:tool}(A2)), which is then used by our model to estimate new viewpoints that meet these criteria. 
In addition to brushing, users can specify a 2D hyperplane over parts of the scene and define the number of locations  to be retrieved for each query (Fig.~\ref{fig:tool}(A3))
This will prompt the model to estimate viewpoints for the desired number of locations that satisfy the brushing constraints in the hyperplane.

\noindent \textbf{Latent map view.} As illustrated in Fig.~\ref{fig:tool}(B), this view presents the model’s second-to-last layer predictions projected in a 2D scatter plot using semi-opaque purple points.
By default, it displays predictions for the entire ground truth test set; when users brush the PCP (Fig.~\ref{fig:tool}(A2)), the view updates to show only the predictions corresponding to the brushed subset. 
In addition, the newly estimated viewpoints produced by the model based on the query constraints from the PCP are shown as fully opaque green points in the scatter plot. 
Users can choose projection type and optionally zoom in and pan the scatter plot (Fig.~\ref{fig:tool}(B1)).

\noindent \textbf{Gallery view.} Users can brush the scatter plot to highlight specific points (Fig.~\ref{fig:tool}(B2)), and the corresponding viewpoints for the brushed green points are then displayed in the gallery view (Fig.~\ref{fig:tool}(C). The gallery view is scrollable (Fig.~\ref{fig:tool}(C1) and upon selecting one of the viewpoints (Fig.~\ref{fig:tool}(C2) from the gallery, it will then update the 3D map view with the corresponding camera perspective.

\noindent \textbf{3D map view.} The 3D map view (Fig.~\ref{fig:tool}(D)) initially displays the entire 3D urban scene and supports zooming, panning, rotation, and tilt. 
Upon selecting a viewpoint (Fig.~\ref{fig:tool}(C2)) from the gallery, the 3D scene updates with the corresponding camera perspective. 
Users can optionally compute the distribution of visibility elements such as sky, tree, or water, from building facades (Fig.~\ref{fig:tool}(D1)) and visualize it over the facade tiles (Fig.~\ref{fig:tool}(D2)). Users can also create queries by brushing the PCP to display specific visibility ranges on facade tiles, which requires enabling the building view query checkbox from the control query panel. 
A color legend (Fig.~\ref{fig:tool}(D3)) indicates the distribution of the selected visibility element visualized in the facades.

\noindent \textbf{Design justifications.} 
We chose PCP (Fig.~\ref{fig:tool}(A)) primarily for its ability to create query constraints through simple brushing (Fig.~\ref{fig:tool}(A2)) along multiple axes, making it well-suited for querying multivariate data. 
Additionally, it effectively visualizes the distribution of ground truth visibility data (e.g., sky, building) in a clear and interpretable manner. 
To reduce visual clutter when many variables are present, we included toggle buttons (Fig.~\ref{fig:tool}(A1)) that allow users to add or remove axes from the plot as needed. To facilitate exploration of how the ground truth relates to the spatial arrangement of model predictions, we chose to use a scatter plot (Fig.~\ref{fig:tool}(B)) with selectable projection methods (Fig.~\ref{fig:tool}(B1)). 
Given the high density of data points and potential visual clutter, we enabled zooming and panning (Fig.~\ref{fig:tool}(B1)) for better navigation.
To address overlapping points, we reduced their opacity. To distinguish the newly generated camera views resulting from PCP queries, we used a different color (green) for these points and kept them fully opaque, as their number is relatively small. To display the generated viewpoints, we used a scrollable gallery layout (Fig.~\ref{fig:tool}(C1)), as the number of views can often exceed the available window space. 
We made the 3D scene fully explorable (Fig.~\ref{fig:tool}(D)), supporting pan, zoom, rotate, and tilt interactions. 
To enable seamless navigation to model-generated viewpoints, we implemented selection of a view from the gallery (Fig.~\ref{fig:tool}(C2)), which automatically updates the 3D scene to that camera perspective. 
Additionally, to support intuitive visibility exploration of elements (e.g., water, sky) from buildings, we used facade tiles (Fig.~\ref{fig:tool}(D2)) to reveal the distribution of visibility through a color scale (Fig.~\ref{fig:tool}(D3)).

\section{Evaluation}

We evaluate our approach through a combination of quantitative and qualitative analyses to assess both the accuracy and performance of the view estimation model, as well as its practical utility in real-world urban analyses.
Quantitatively, we measure the accuracy of the predicted view-based distributions against ground-truth data generated via rasterization (Section~\ref{sec:accuracy}), and assess computational performance in terms of inference speed (Section~\ref{sec:performance}).
Qualitatively, we demonstrate our approach's effectiveness across three use cases (Section~\ref{sec:cases}) \highlight{and feedback from experts (Section~\ref{sec:feedback}).}

\subsection{Datasets}
\highlight{For all the experiments and use cases, we relied on a publicly available 3D urban dataset with mesh-based representations. In particular, mesh data was downloaded from Geopipe~\cite{mitchell2025geopipe}. The dataset includes a region of Manhattan, New York City, where the density of buildings, infrastructure, and vegetation present diverse occlusion and visibility conditions.
Our testbed covered 3 $km^2$, containing 3,488 building structures and 113,699 vertices.
}

\figEvaluationAccuracy

\subsection{Quantitative experiments}

\subsubsection{Accuracy analysis}
\label{sec:accuracy}

\myparagraph{Evaluation of prediction accuracy}
To evaluate the prediction accuracy of our model, we designed an experiment involving synthetic thematic data, simulating common material-based classes observed in urban environments.
We generated six scenarios, each with a different balance between two thematic classes: \emph{brick} and \emph{glass}. For each scenario, buildings in the scene are randomly assigned one of the two classes according to the specified ratio (e.g., 90\% brick and 10\% glass Fig.~\ref{fig:evaluation_accuracy}).
This evaluation aimed to test the model's ability to generalize under varying thematic distributions.

We trained the model for 100 epochs to predict, for each view, the distribution of visible thematic content. Specifically, the percentage of brick versus glass facades observed from that viewpoint.
To compute the error, we divided the urban scene into square regions (each roughly corresponding to the size of a city block), and each region was evaluated from 8 uniformly sampled directions. We computed the error between the predicted and ground-truth distributions for each directional view.
Table~\ref{tab:region_with_small_error} shows the percentage of regions where the prediction error for each class is below 10\%. The results show high accuracy across a wide range of class balances. For example, in the most imbalanced scenario (99\% brick, 1\% glass), the model still achieves 82\% accuracy on brick regions and 99\% on glass regions.
Even in the balanced case, performance remains strong, with 84.84\% of brick regions and 83.61\% of glass regions falling below the 10\% error threshold.



\begin{table}[tb]
\caption{Percentage of regions with under 10\% error for thematic data with two classes across varying class balance distributions.}
\label{tab:region_with_small_error}
\scriptsize
\centering
\begin{tabular}{lcccccc}
\toprule
\textbf{Brick / Glass Ratio} & 99/1 & 90/10 & 80/20 & 70/30 & 60/40 & 50/50 \\
\midrule
\textbf{Brick regions (\%)}  & 82.88 & 78.26 & 79.30 & 81.46 & 78.24 & 84.84 \\
\textbf{Glass regions (\%)}  & 99.13 & 92.43 & 87.49 & 84.03 & 80.83 & 83.61 \\
\bottomrule
\end{tabular}
\end{table}

\myparagraph{Comparison with small-footprint ML models}
To evaluate the accuracy of our model relative to other lightweight machine learning models, we conducted a comparative study using a large dataset of labeled views.
The task involved predicting the distribution of seven view indices (building, water, road, sidewalk, surface, tree, sky) from the location and view direction of each input.
The training set consisted of 63,696 views, and the held-out test set included 15,000 views.
To assess performance under different training data densities, we trained our model on subsets of increasing sizes of the full training set (10\%, 20\%, 30\%, 80\%, and 100\%) and evaluated the root mean squared error (RMSE) on the test set.
We conducted the same experiments using two lightweight models: k-nearest neighbors (KNN) with varying numbers of neighbors, as well as a random forest regressor. Each model was trained on identical data splits, and all results were evaluated after 100 epochs.

As shown in 
Fig.~\ref{fig:evaluation_comparison}, our model consistently outperforms all other models across training set sizes, particularly in low data scenarios.
For example, with only 10\% of the training data (6,000), our model achieved an RMSE of 0.078, compared to 0.103 for both the 2- and 5-neighbor KNN models, and 0.112 for the random forest.
As the training set size increases, our model continues to improve, achieving 0.046 RMSE on the full dataset.


\figEvaluationComparison

\subsubsection{Computational efficiency analysis}
\label{sec:performance}

We evaluate the computational performance of our method by measuring the number of views it can infer per second.
All performance experiments were conducted on a MacBook Pro M1.
Our model achieves an average inference rate of approximately 4,000 views/second, considering a batch size of 1.
However, due to the compactness of our model (it only occupies 2.4 MB of memory), much larger batch sizes (exceeding 1,000) are feasible. The effective throughput scales to approximately 4 million views/second in such configurations.

To contextualize these numbers, consider a building with a height of 1,000 feet and a ground-level footprint of $100 \times 100$ square feet. Its four facades can be subdivided into 1,000 patches, each measuring $20\times20$ square feet. Estimating the visibility of each patch from five different viewpoints would require 5,000 view evaluations. Our model can complete this task in approximately 1 to 2 seconds (using a batch size of 1).
In contrast, a traditional approach based on rasterization would require over 160 seconds, assuming a rendering rate of 30 frames per second.

\subsection{Use cases}
\label{sec:cases}

We demonstrate the capabilities of our model through a series of use cases, illustrating how urban experts can leverage the view estimation model to explore and analyze view-based thematic information.
The first use case highlights how our model can be used for view-based exploration of thematic data.
The second use case illustrates how users can define desired view characteristics and use inverse queries to identify viewpoints satisfying custom view constraints.
Finally, the third use case focuses on analyzing views from buildings, showing how facade-level viewpoint estimation can support visibility analysis.



\subsubsection{Exploration of thematic data using direct queries}

In this use case, we demonstrate how our model and system can support the view-based exploration of thematic data.
The thematic data is derived from the visibility of buildings towards their surrounding environment. Each building is then annotated with view indices computed from a dense sampling of cameras placed along its facade surfaces. 
%

%
In a concrete example, an urban analyst might be interested in identifying buildings whose facades have high visibility to trees. The user selects the upper range of the ``tree'' axis in the query panel, filtering down to views towards buildings that satisfy such a condition.
These filtered views are then projected into the 2D latent map, which encodes similarity across views.
This view allows users to identify clusters or outliers. Selecting a region in this latent space updates the results gallery, which displays sample views from the selected buildings.
These view thumbnails validate the thematic metrics. Clicking a view in the gallery centers the camera on the corresponding viewpoint, allowing users to inspect its geometric surroundings and assess visibility conditions (Fig.~\ref{fig:teaser}).


This use case illustrates how our system supports exploration of thematic data without the need for manual 3D navigation. By combining multivariate filtering and the image gallery, users can analyze visibility patterns across the city and ground their interpretations in quantitative data and representative views.


\figUseCaseInverseQueries

\subsubsection{Viewpoint discovery using inverse queries}

This use case demonstrates how our model supports \emph{inverse queries} to locate viewpoints that match user-defined goals.
Users can specify desired characteristics and allow the system to search for camera positions and directions that yield views satisfying these conditions.

In this case, the user begins by defining the characteristics of interest through the query control panel. In one scenario, they may specify a custom perception metric, built as a weighted combination of values such as openness, visibility of green space, or proximity to landmarks. Alternatively, the user may prioritize high sky exposure and green space visibility while minimizing building density. This could reflect, for instance, a goal to identify locations in the city that feel open, natural, and comfortable for pedestrians~\cite{ma_2021_perception}.
Once a view distribution has been defined, users specify a spatial region over which the query should be applied. This is also done in the query control panel. The user can fine-tune the selected surface directly in the 3D environment if needed.

With the view distribution and spatial domain set, the system initiates the inverse query. It evaluates potential viewpoints within the defined region and returns those whose predicted view distributions most closely match the user's specification.
The top-matching views are presented in the gallery, allowing users to compare candidate perspectives side by side (Fig.~\ref{fig:use_case_inverse_queries}).
This use case illustrates how the system supports goal-driven exploration through inverse queries from desired perceptual conditions.


\figUseCaseFromBuildings

\subsubsection{Quantifying visibility from buildings}

This use case focuses on analyzing what is possible to see from the perspective of buildings. Each view is associated with a camera placed on the building's surface, and is evaluated using the same view indices described in the previous use cases (e.g., sky exposure, water visibility, green space). This enables visibility analysis with applications in real estate~\cite{kaimaris_real_estate_views_2017}, architectural evaluation~\cite{varoudis_architecture_visibility_2014}, and environmental comfort studies~\cite{deluca_comfort_2019}.

The workflow begins by selecting the building analysis mode in the query control panel, switching the system into a mode where views are computed from building facades rather than towards them. The user then navigates to a building of interest using the 3D exploration map. For example, an analyst might focus on a proposed residential tower near a riverfront.
Once the building is selected, the user chooses a view theme to analyze. This might include, for instance, computing how much of the view from each point on the building's facade includes visible water, greenery, or sky.
In the query panel, the user can define these thematic goals. The system then evaluates views from multiple sample points on the building's surface, generating a set of view distributions.
These are visualized directly on the facade in the 3D component, with colors reflecting the chosen view indices (Fig.~\ref{fig:use_case_from_buildings}).
Users can switch between different thematic layers to compare how various types of visibility play out across the facade.
In parallel, the gallery is populated with individual views from the selected building, sorted by the currently active view index. 
The gallery allows for a qualitative inspection of the views from the building. Click on any thumbnail links to return to their corresponding position on the facade.
This case highlights how the system and model can support building-scale analysis by computing visibility indices, which is especially important to support decisions around design~\cite{nasar_house_design_1994,Yap2023_epb_semantic_networks}, valuation~\cite{lu_house_pricing_2018}, and regulation~\cite{BRASEBIN_regulations_2018}.

\subsection{Experts' feedback}
\label{sec:feedback}

\highlight{
To complement our use case demonstrations, we presented the tool to a group of domain experts and collected their impressions. 
We conducted six 30-minute, semi-structured interviews, during which we explained the goal of our model and demonstrated it through the use cases described in Section~\ref{sec:cases}.
The experts then reflected on potential applications, limitations, and extensions.
We gathered feedback from six experts:
E1 is a doctoral student with 3 years of experience in geography and environmental studies. 
E2 is a postdoctoral researcher with 7 years of experience in design and engineering.
E3 is a doctoral candidate with 7 years of experience in design and engineering.
E4 is an assistant professor with 12 years of experience in civil engineering.
E5 is a doctoral candidate with 15 years of experience in architecture.
E6 is an assistant professor with 10 years of experience in urban planning.

%

Regarding \textbf{domain applications}, E1 mentioned that our ``method in considering water, trees, and greenery in one composite or thematic version is a breakthrough in visibility analysis.''
%
Regarding applicability for planning, E2 mentioned our method's usefulness for what-if analyses, mentioning that ``Remove the trees, what views are affected? What-if scenarios (...) can be a good asset for planners.''
E3 mentioned that ``it's super relevant for urban planning, and it's something that I’ve never seen before, so I think it’s a really exciting development,'' further mentioning that the variables used (open space, water) can ``really play an important role in shaping people's mental health.''
Multiple experts also mentioned the value of our contributions for real estate.
E2 mentioned that ``You could then filter, okay, give me the view that is common for high-income people, or, given an income, what views can I afford?''
E3 mentioned that ``One of the use cases that could be really relevant would be real estate and finance.'' He continued saying that ``currently the approaches that we have in real estate are not that good (...) it's mostly pre-computed and they are not able to simulate new viewpoints in a dynamic way like you have done.''

Regarding \textbf{interactivity}, E3 was particularly impressed by the capability of composing indices. He highlighted that ``since you have these diverse semantic metrics, you can combine them to find a composite index of how people perceive space,'' continuing that ``you can simulate many different kinds of indicators that help people understand space better.''
E5 stressed that ``there is a lot of merit to making these workflows faster (...) making them real-time and interactive to support open-ended exploration.''
E5 further highlighted the model's potential for navigation, mentioning that she could see a use case for finding ``the most comfortable route to walk from A to B,'' where users choose to maximize ``exposure to amenities or open streets or sky view factor, greenery, etc.''
She further mentioned that she sees value in not just viewing recommendations but ``how can users start viewing it as a way to explore the city, or a route, for instance.''

Experts also highlighted potential directions for \textbf{extensions}.
E5 highlighted that in the future we could incorporate ``additional layers such as urban radiation with watts per square meter so that users could query for the most comfortable spots.''
Reflecting on his practice, E1 mentioned that it would be interesting to have users select ``landmarks so that views towards them could be analyzed.''
E3 mentioned that he would be interested in using our model to create \emph{fingerprints} for buildings, defined as ``average of the thematic information of each building,'' so that they could be used for ``predicting building function.''
E6, whose work focuses on sidewalk accessibility, suggested that a valuable extension could be to allow users to ``specify the level of analysis.'' In her case, this would mean evaluating views specifically at the street level, with the goal of constructing an index of walkability that incorporates factors such as the visibility of trees and landmarks.
Reinforcing E1's earlier point, E6 also emphasized the importance of enabling users to define landmarks directly within the interface, which she considered a particularly useful addition for walkability-related assessments.
Regarding the interface, E6 acknowledged that her first impression was one of ``visual overload,'' given the number and variety of components presented simultaneously, but found it intuitive after brief exploration. When asked about the latent space, she noted that ``its usefulness became clearer through interaction.''
}
\section{Conclusion and Discussion}
In this paper, we introduced a view-based 3D urban data exploration technique.
At the core of our approach, we developed a neural field model for visibility estimation. We detailed the training pipeline and how to achieve the encoding of physical and thematic data visible from locations throughout the city.
We further enhanced the model to include two extensions, one for the input and one for the output. The input regularization module enables a targeted analysis of city subspaces, such as planes hovering over street networks. The output customization module allows the user to define further abstract view indices
as algebraic expressions of the default thematic features.
As building blocks for data exploration, we proposed two types of operations on top of our visibility estimator: direct and inverse queries. Through direct queries, the user can compute the visibility thematic for a large number of input locations. At the same time, inverse queries allow the user to find spatial locations for which a custom set of visibility constraints is satisfied. 
%
We evaluated the performance of our visibility estimator through accuracy tests, comparisons with other small-footprint machine learning models and speed tests against a ground truth scene renderer.
As for practical applications, we showcased three urban use cases in which our methodology can be employed. First, we explored thematic data associated with building facades. We showed how to find areas satisfying user-specified visibility constraints on a large number of building facades. Second, we used surface parametrization and gradient-based inverse queries to find views alongside a custom-selected street. Through the exploration of the latent space generated by our model, we were able to get further visibility into insights. Thus, we discovered view sub-patterns within the returned locations. Finally, we estimated views from the perspective of buildings and compared the results based on the location of each building facade.    
\highlight{During our meeting with experts, they emphasized our contribution's relevance across urban planning, real estate, and environmental applications, while also envisioning extensions such as landmark and radiation analyses.}

\myparagraph{Limitations}
We identify a few limitations in our technique and further propose potential solutions.
Since our model is trained on a representation of the 3D physical world, there is a natural imbalance between the existing visibility classes.
For example, buildings will occupy a much higher percentage of the view space than street furniture.
This imbalance leads to regression dilution for the underrepresented classes, in which the model tends to predict lower visibility values for these classes even in areas where they are more present.
One solution for this problem would be to use log scales to quantify the thematic data.
This scaling would be consistent with the human perceived view impact, which increases logarithmically with respect to the vision stimuli, as mentioned in the \textit{Weber–Fechner Law}~\cite{DEHAENE_Weber_Fechner_2003}. 
Alternatively, we could use a class-weighted loss, which would dynamically adapt its weights based on the spatial location of the input.
Another limitation of our approach is that our model's precision in visibility could be reduced when modeling larger spacial extents and/or more geometrically detailed areas.
One possible solution for this is to train multiple models to cover specific city areas, allowing for higher precision in particular areas.
This idea was shown to have a good precision for applications targeting the rendering of urban environments~\cite{partha2023neural}.  
%
%
\highlight{Finally, our model assumes mesh-based urban representations, which may limit its direct applicability to other formats such as point clouds or textured models.}

\myparagraph{Future work}
We identified several opportunities for extending our current work.
\highlight{First, while our approach is in principle agnostic to the specific urban or suburban setting, our evaluation has focused on environments where building geometry plays a central role. Extending it to less dense and rural settings (e.g., landscapes with minimal built structures) is an interesting avenue for future work, particularly for applications in landscape analysis~\cite{swietek2023visualcapital} and rural spatial planning~\cite{fujiwara2025voxcity}.}
We also plan to integrate a number of additional types of thematic data into our analysis, such as sunlight access, flooding exposure, noise propagation, and traffic intensity.
Second, we plan to implement this technique as part of our growing open source toolkit for urban data visualization~\cite{moreira2024utk,moreira2025curio}.
%
%
Third, a direct path towards improving the precision of our model would be dividing the scene into regions and training individual specialized models for each area.
Furthermore, the model could be extended to support ``what-if'' scenarios, for instance, by simulating the view impact of a new construction through regenerating training data in its proximity and fine-tuning the model on the fly with this augmented dataset.
\highlight{Finally, a promising direction for future work is a dedicated study with urban experts to evaluate how latent space representations of views can support domain-specific analysis and decision-making.}

\acknowledgments{%
We thank the experts who participated in our evaluation and the reviewers for their constructive feedback. This work was supported by the U.S. National Science Foundation (Awards \#2320261,
\#2330565, \#2411223), IDOT (TS-22-340), CNPq (311425/2023-2) and conducted as part of the Open-Source Cyberinfrastructure for Urban Computing (OSCUR) project.
}

\bibliographystyle{abbrv-doi-hyperref}

\bibliography{main}

\end{document}